\documentclass[journal]{IEEEtran}

\usepackage{amsmath,amsfonts}
\usepackage{graphicx}
\usepackage{cite}

\usepackage{subcaption}
\usepackage{tabularx}
\usepackage{lipsum}

\usepackage{gensymb}
\usepackage{algorithm}
\usepackage{algorithmic}
\usepackage{tabularx}
\usepackage{makecell}

\begin{document}

\title{Tracking Wildfire Assets with Commodity RFID and Gaussian Process Modeling}

\author{John Hateley, Sriram Narasimhan, Omid Abari%
\thanks{Hateley and Narasimhan are with the Department of Mechanical \& Aerospace Engineering at the University of California, Los Angeles and Abari is with the Department of Electrical and Computer Engineering at the University of California, Los Angeles (e-mail: jmhateley,snarasim@g.ucla.edu, omid@cs.ucla.edu).}%
\thanks{This is a pre-print version of the paper that has been accepted by IEEE Journal of Radio Frequency Identification, without any of their recommended changes. Some of these changes include corrected figures, additional context and figures. The full version with added figures and text can be found under the same title here https://ieeexplore.ieee.org/document/11298452.Personal use of this material is permitted. Permission from IEEE must be obtained for all other uses, including reprinting/republishing this material for advertising or promotional purposes, creating new collective works, for resale or redistribution to servers or lists, or reuse of any copyrighted component of this work in other works.}
}

\maketitle

\begin{abstract}
 This paper presents a novel, cost-effective, and scalable approach to track numerous assets distributed in forested environments using commodity Radio Frequency Identification (RFID) targeting wildfire response applications. Commodity RFID systems suffer from poor tag localization when dispersed in forested environments due to signal attenuation, multi-path effects and environmental variability. Current methods to address this issue via fingerprinting rely on dispersing tags at known locations {\em a priori}.  In this paper, we address the case when it is not possible to tag known locations and show that it is possible to localize tags to accuracies comparable to global positioning systems (GPS) without such a constraint. For this, we propose Gaussian Process to model various environments solely based on RF signal response signatures and without the aid of additional sensors such as global positioning GPS or cameras, and match an unknown RF to the closest match in a model dictionary. We utilize a new weighted log-likelihood method to associate an unknown environment with the closest environment in a dictionary of previously modeled environments, which is a crucial step in being able to use our approach. Our results show that it is possible to achieve localization accuracies of the order of GPS, but with passive commodity RFID, which will allow the tracking of dozens of wildfire assets within the vicinity of mobile readers at-a-time simultaneously, does not require known positions to be tagged {\em a priori}, and can achieve localization at a fraction of the cost compared to GPS. 
 
\end{abstract}


\subsection{Problem}
According to various firefighting agencies we interviewed, including the California Department of Forestry and Fire Protection (CAL FIRE), the US Forest Service, and the Mountains Recreation and Conservation Authority (MRCA), the number of firefighters deployed during a wildfire can range from 25 to over 5,000, depending on the fire's scale. Firefighters must carry numerous pieces of equipment and tools to perform their duties effectively and often need to move hundreds, or even thousands, of meters from their vehicles. This wide range of deployment and the potential inclusion of thousands of personnel with their own equipment can result in a substantial dispersion of assets, including both human resources and various tools. In such situations, management teams require near real-time situational awareness of their assets—both personnel and equipment—to ensure resources are utilized efficiently and firefighter safety is maintained.

Currently, Global Positioning Systems (GPS) are used to track high-value assets, allowing for convenient monitoring of their locations to within 3-5 meters under ideal weather and terrain conditions \cite{gps_1}. However, constraints such as limited power, lack of supporting infrastructure (e.g., cellular connectivity for GPS-equipped phones), budgetary considerations, and terrain effects make GPS and GPS-enabled devices practical for deployment only on a few high-value assets. Other technologies such as Bluetooth and Ultra Wide Band (UWB) sensor tracking devices can cost between \$20 -- 100 per unit, limiting the number of assets that can be tracked using them cost-effectively. Furthermore, power constraints such as battery life (typically one year or less) adds additional practical constraints. Conventional methods of communication such as verbal over the radio provide very sparse information and often fail to deliver real-time situational awareness to the management team regarding the assets deployed for firefighting. Hence, there is a practical need to develop scalable and cost-effective solutions to track wildfire assets.



\begin{figure}[t!]

    \centering
        \includegraphics[width=0.48\textwidth]{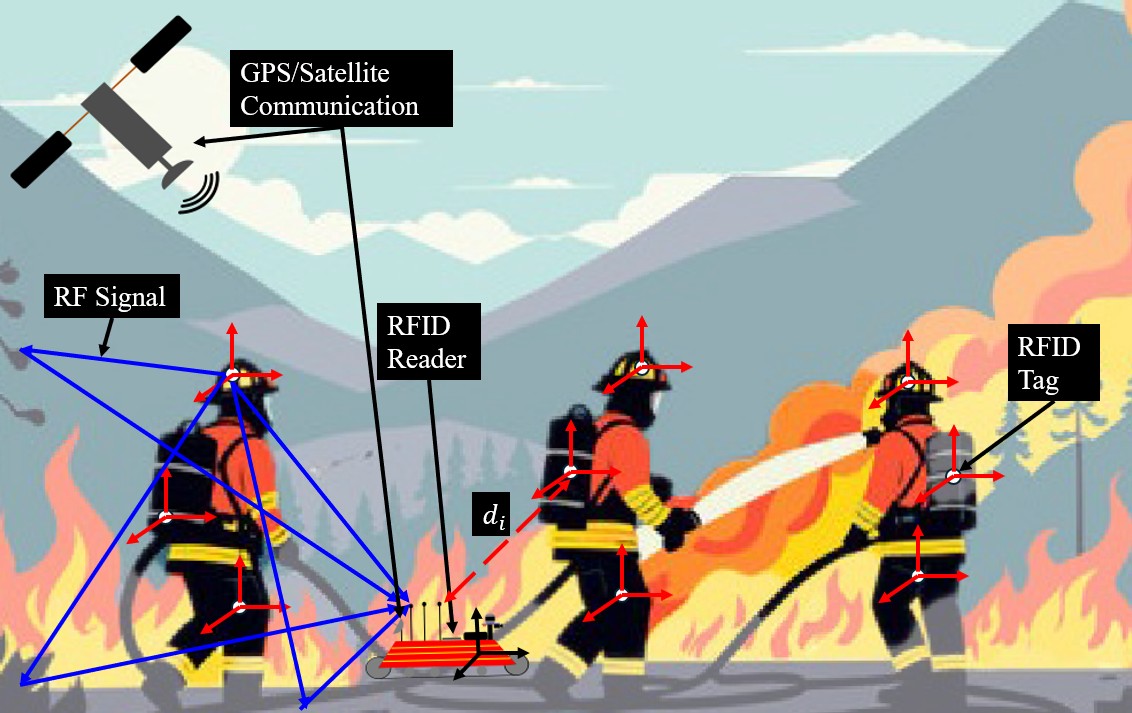}  
        \caption{Proposed concept of commodity RFID to track wildfire assets where RFID tags are distributed across assets and are tracked using a system of readers mounted on mobile vehicles.}
        \label{fire_pic_concept}  
    \label{fig:main_fig}
\end{figure}

\subsection{Approach}
We propose to combine geo-located high-value mobile assets—such as firefighters and vehicles equipped with localization capability—with ubiquitous passive RFID sensing to overcome the practical cost and power challenges of tracking multiple assets at-a-time in the vicinity of the reader. When implemented in a distributed setting through the use of a number of mobile RFID readers, commodity RFID technology can be cheap\textemdash passive RFID tags cost around a \$1-5 or less each, with one reader capable of supporting 100 tags\textemdash and require no power. This makes them easily scalable and capable of tracking hundreds of assets at a fraction of the cost of GPS technology. When deployed at ground level, RFID can operate effectively under dense canopies and in GPS-denied environments. However, the primary limitation of RFID technology lies in its ranging accuracy, which depends on various factors, including attenuation, multi-path effects, and environmental factors\cite{tag_10}.

Therefore for RFID technology to be effective in this application, we first need to demonstrate that passive RFID technology can achieve spatial positioning errors comparable to those of conventional GPS and commonly used active sensors (0.5-5 meters) at distances ranging from a few meters to tens of meters from RFID readers, particularly in forested environments \cite{gps_1}. Previous studies have shown that inertial navigation systems using GPS and simultaneous localization and mapping (SLAM) can achieve sub-meter level spatial positioning \cite{slam1}. Thus, by integrating RFID technology with GPS and SLAM-based systems, we can significantly enhance situational awareness for management teams during active wildfires, enabling them to quickly and effectively locate and distinguish between deployed assets.

\subsection{Contributions}
This paper distinguishes itself from other sensor localization studies in several key ways. Obviating the need for {\em a priori} tagged ground truth locations in an environment, it addresses the signal attenuation, multipath and environmental effects through a powerful machine learning approach—specifically, Gaussian Process (GP) regression and then compares this technique to the K-th Nearest Neighbor Algorithm\cite{KNN_1,KNN_2}, a localization technique that is frequently used as a basis for RFID tracking \cite{tag_9,tag_23,tag_24}. Unlike previous methods that rely on geo-spatial coordinates and physical characteristics to identify the environment, our approach solely utilizes the signal response to achieve the necessary accuracy, making it more practical for deployment in this application. Extensive experiments are conducted in six forested environments to demonstrate the framework. To our knowledge, this is the first work to demonstrate that commercial RFID technology can achieve sub-meter localization accuracy in wildfire-prone environments by leveraging data-driven GP regression modeling. We further show how to match an RF environment to the closest entry in a model dictionary using only signal attributes—an essential step in achieving reliable performance. We believe that the ideas proposed here is scalable when implemented in a distributed setting where RFID readers are mounted on mobile assets such as firefighters or vehicles and the cluster of tags mounted on assets in the vicinity of the readers are tracked using mobile readers.


This paper is organized as follows. In Section \ref{lit_rev}, we discuss work related to RFID sensor localization and GP regression. We detail how experiments were conducted and the modeling framework used for sensor localization and environmental identification in Section \ref{method}, with the specifics for how the data was collected in Section \ref{implementation}. The results of our model and comparison to a well-established localization technique are then discussed in Section \ref{results}, with the key conclusions from this study and limitations presented in Section \ref{conc}.

\section{Background and Related Work}\label{lit_rev}


\subsection{RFID Localization}

A passive RFID system consists of a reader and a passive tag, which relies on the reader’s high-power RF signal for activation. The tag responds via backscatter using ON-OFF keying modulation to transmit its ID. In addition to the ID, RFID readers provide signal features such as phase ($\phi$), frequency ($f$), time of reception, and RSSI, which can be used for localization via techniques like Phase of Arrival and RSSI ranging, as seen in Equations \ref{phase_equation} and \ref{RSSI_equation}. These equations incorporate reference measurements ($\phi_0$, $RSSI_{ref}$), the path loss exponent ($n$), signal wavelength ($\lambda$), and system-specific error terms ($\epsilon_1$, $\epsilon_2$) to localize the tags based on antenna positions \cite{acm1,acm2}.


\begin{equation}
    \phi[i] = \phi_0 + \frac{4 \pi d}{\lambda} + \epsilon_1    
    \label{phase_equation}
\end{equation}

\begin{equation}
    RSSI[i] = RSSI_{ref} - 10n\log(d) + \epsilon_2
    \label{RSSI_equation}
\end{equation}

The antenna's orientation to the tag influences the Received Signal Strength Indicator (RSSI) and phase equations. Therefore, it is preferable to use a ranging method that does not depend on the alignment between the tags and the antenna. One such method is the Phase Difference of Arrival (PDOA), as described in Equation \ref{phase_diff_equation}.

\begin{equation}
    \Delta\phi[i] = \frac{4 \pi \Delta f d}{c} + \epsilon_3
    \label{phase_diff_equation}
\end{equation}

$\Delta\phi[i]$ is obtained by subtracting the phase values, $\phi_1[i]$ and  $\phi_2[i]$ at two adjacent frequencies, $f_1$ and $f_2$, called the phase difference. Assuming orientation and various system offsets do not change, the effects of the environment remain and are represented by $\epsilon_3$. Through these equations, the tag's distance from the antenna can be combined with various machine learning techniques, such as neural networks or environmental fingerprinting, to determine the tag's coordinate $(x_t)$ by expressing it as the solution to a nonlinear optimization problem, such as the one shown in Equation \ref{tag optimization eq}: 

\begin{equation}
    \min_{x_t}\sum_{i = 1}^{N} (d_{i} - \lVert x_{a_{i}} - x_t \rVert )^2
    \label{tag optimization eq}
\end{equation}
where, $d_i$ is the calculated range and $x_{a_i}$ represents the antenna's position\cite{dis_equation}.

 RFID tags have seen application in numerous localization scenarios, such as tracking marked assets inside and outside of homes and tracking deployed mobile robots \cite{acm1,acm2,ieee_tag_1}. For instance, in this study \cite{tag_1}, the positions of a uniform linear array of tags were tracked using a moving antenna. In these studies \cite{tag_6,tag_10,tag_18}, RFID tags were distributed in an environment to act as landmarks to locate mobile drones. But, since these tags operate on short-range communication, obstructions in the environment, such as walls, pillars, or even weather, can cause the signal to reflect or distort \cite{tag_14}. This distortion causes fluctuations in the RF signal and reduces the localization accuracy. To overcome this issue, filters are commonly used to remove noise from the signal to allow the system to determine the tag's true distance from the antenna \cite{tag_15}. However, these filters can be complex and require careful tuning to ensure localization accuracy. As a result, an alternative to using filters is to model how the RFID signals behave in an environment, known as Environmental Fingerprinting. In RFID applications, fingerprinting involves determining the incoming RFID signal from various tags dispersed in an environment with known locations (ground truths). This helps quantify the influence of the environment on the radio waves while avoiding an in-depth analysis of the environment itself \cite{tag_2}. This approach to tag localization has been utilized successfully in various studies, such as in \cite{tag_9} where fingerprinting was used to train a Convolutional Neural Network or to track a marked robot or assets as in \cite{tag_17} and \cite{tag_24}. Fingerprinting requires knowledge of the environment and signal behavior at specific positions a priori and cannot capture the changes over time \cite{tag_23}, which may be limiting in some applications such as tracking assets in the wild. Although these algorithms have been proven to be successful in tracking RFID tags, these models require specific environmental information in order to operate properly. However, due to spontaneity of wildfires and the need for rapid response, it is not possible to thoroughly map the signal response in the specific environment the tag will operate in, thus preventing the use of these algorithms in this application.

\subsection{Gaussian Process Modeling} \label{gp_review}

Gaussian Process (GP) regression is a statistical modeling technique, which can be viewed as a generalization of a multivariate Gaussian distribution to an infinite dimension. Assume that the outputs in $\mathbf{y}$ are generated by inputs $\mathbf{x}$ through a general nonlinear regression Equation \ref{nonlin_reg}:
\begin{equation}
    {\bf y}=f({\bf x})+\epsilon
    \label{nonlin_reg}
\end{equation}
where ${\bf \epsilon}$ is noise sampled from a standard normal distribution. In GP regression, the function $f({\bf x})$ is a distribution over functions, defined by:
\begin{equation}
    f({\bf x}) \sim \mathcal{GP}(m({\bf x}), \mathcal{K}({\bf x}_1, {\bf x}_2))
   \label{GP_fund}
\end{equation}
through the use of a covariance function ($\mathcal{K}({\bf x}_1,{\bf x}_2)$) and a mean function ($m({\bf x})$), as shown in Equation \ref{GP_fund}\cite{Rasmussen_Williams_2006}. 
The covariance function ($\mathcal{K}({\bf x}_1,{\bf x}_2)$) models the dependence between the function values at inputs ${\bf x}_1$ and ${\bf x}_2$ \cite{Rasmussen_Williams_2006}. The functional form of the covariance function is determined by the {\em kernel} $\mathcal{K}$ and is based on the data itself, such as decaying values with distance. While GPs involve distributions over functions, in practice, the function values are computed only on finite points, which makes their application practically feasible. The posterior mean and covariance of a GP involve calculating covariance matrices that scale with the training data size. However, for applications such as the current one with training data sets in the hundreds, this is not an issue. In many applications, the mean function is assumed to be 0 or achieved by subtracting the mean from observations, which further simplifies the posterior estimation \cite{Rasmussen_Williams_2006}. The mathematical details of GPs have been extensively studied and hence not repeated here for the sake of brevity; readers are referred to seminal works such as this \cite{Rasmussen_Williams_2006} for details.

Neural Networks and their variations have been used to localize RFID tags, e.g., \cite{neural_1,neural_2,neural_3}. However, GPs are advantageous for this application as they can model uncertainty explicitly which is crucial when dealing with sparse data and incomplete environmental model dictionaries. GP models have been studied in the literature in several applications, including robotics and weather prediction. In \cite{GP_2} and \cite{GP_10}, GP Models were used to predict temperature and weather, based on sparse sensor measurements, with \cite{GP_9} discussing how the different models affect the accuracy of the maps. In \cite{GP_1} GP model was used for structural mapping and robot path planning, and in \cite{GP_7} for drone tracking using radar fusion. A comprehensive review of GP cannot be attempted in this paper due to the large volume of applications; suffice to say, despite GP's popularity and broad applications, GPs have not been used previously in the current context namely, wildfire asset tracking using RF signal attributes.

\subsection{Literature Gaps}

The primary gap in the literature is the application of GP models for RFID tag localization and environmental classification, particularly in scenarios where physical features or geo-spatial coordinates are unavailable. While GP models have been utilized for data prediction, they typically rely on prior knowledge of the environment. However, this information is often lacking in wildfire situations, as firefighters may not be familiar with the terrain until they arrive at the fire. Furthermore, forests do not feature a uniform terrain, necessitating that model selection be localized and based on the immediate surroundings. Consequently, the deployed system must classify its environment to select the appropriate model. Existing studies \cite{env_id_1, env_id_2, env_id_4} have attempted environmental modeling by relying on physical attributes of space (e.g., walls, material composition) and employing sensors such as infrared cameras or ultra-wideband technology. However, these methods do not utilize RFID technology, perhaps the most cost-effective passive solution for large-scale applications. Additionally, obstructions such as trees can hinder visual sensors' ability to extract environmental features. Therefore, the system must depend solely on the signal itself, without the benefit of geo-spatial coordinates or identifiable physical features, to classify the environment and determine the suitable model. This collection of challenges has not been adequately addressed in previous research, signaling a new area for inquiry.

 \section{Methodology}\label{method}

In this section we present the theoretical background of our GP Model and how it used to predict range estimates. We then present how the appropriate models were selected using a weighted log likelihood.

 \subsection{GP Model} \label{GP}
 This section describes the procedure to select state points, kernel, and mean function for GP modeling. Since each environment is unique such as obstructions, their distribution of nonlinearities and noise will also need to be accounted for. Therefore, each environment in our dictionary has an associated GP Model with its associated attributes. 
 
\subsubsection{Model Properties and Generation} \label{state_point}
The first step in GP modeling is to define the inputs and outputs (state points), which are application-specific and will be represented by $\mathcal{X}$ and $\mathcal{Y}$, respectively. As shown in Equation \ref{states}, the outputs $\mathcal{Y}$ contain the tag distances from the antenna, and the inputs $\mathcal{X}$ contain the measured phase differences, which are given by,

\begin{equation}
[\{\mathcal{X}\},~\{\mathcal{Y}\}]=[\{\Delta \phi_1, \Delta \phi_2, \ldots, \Delta \phi_n\},~\{d_1, d_2, \ldots, d_n\}]. 
\label{states}
\end{equation}

Suppose the inputs and outputs are separated into a training set of observations given by $\mathcal{O}_t=\{X_t,~Y_t \}$ and we want to make predictions for inputs $X^*$. Then, it can be shown \cite{Rasmussen_Williams_2006} that the conditional distribution $p(f^*|X_t,Y_t,X^*)$ has the covariance
\begin{equation}
     K(X^*, X^*) - K(X^*, X_t) \left[K(X_t, X_t) + \sigma_n^2 I\right]^{-1} K(X_t, X^*)
     \label{variance}
 \end{equation}
and mean,
\begin{equation}
    K(X^*,X_t)\{K(X_t,X_t)+\sigma_n^2I\}^{-1}Y_t
\end{equation}
assuming the provided data has a 0 mean distribution.

The kernel function $\mathcal{K}$ models the relationship between inputs, and in this application, the $\Delta\phi$ values can have a slight relative difference between them. A natural choice for the kernel is the radial basis function shown in Equation \ref{basis 1},

\begin{equation}
    K(x_1,x_2) = exp \left(-\frac{\lVert x_1 - x_2 \rVert^2}{2\sigma^2}\right)
    \label{basis 1}
\end{equation}

which relates $X_1$ and $X_2$ through their proximity to each other and amplifies the effect of this difference by squaring it \cite{Rasmussen_Williams_2006}. Here, $\sigma$ is the length scale associated with the kernel. 

 The next step is determining the mean function ($m(x)$).This function should capture the {\em ideal} linear relationship between $\mathcal{X}$ and $\mathcal{Y}$, as well as the non-linearities and errors that occur as a result of environmental interference and system noise, as shown in Equation \ref{incoming_phase_signal}: 

 \begin{equation}
     \Delta\phi[i] = \Delta\phi(d) + \mathcal{N}(0,\sigma)+\epsilon.
     \label{incoming_phase_signal}
 \end{equation}

However, due to the inconsistencies in the scale and the appearance of the non-linearities, generating a mean function that incorporates all of these attributes poses a significant challenge, hence the use of a GP Model surrogate. By utilizing the known linear relationship between distance and phase difference and incorporating it with the covariance matrices, as shown in Equation \ref{predic_mean}, 

 
\begin{equation}
    f^* = m(X^*) + K(X^*,X_t)\{K(X_t,X_t)+\sigma_n^2I\}^{-1}(Y_t - m(X_t)) 
    \label{predic_mean}
\end{equation}

the GP model can be used to predict $Y^*$ associated with input data $x \in X^*$ \cite{Rasmussen_Williams_2006}. This method of range prediction is employed in this paper to estimate the distance of the tags from the antenna.

 \subsection{Environment Selection from a GP Model Dictionary}\label{env_id_method}

Central to our method is the ability to associate an environment to the closest match in a model dictionary. This association is accomplished purely based on the signal characteristics observed at some locations in the test area instead of using visual or other physical features. To enable such an association, a dictionary of GP models is created from typical environments and included in the set $\mathcal{M}=\{\mathcal{M}_1,~{\mathcal{M}_2},\cdots,\mathcal{M}_l\} $, where $l$ is the number of environments in the dictionary. Once the models are generated, each of the models are used to generate a set of predictions based their individual parameters and the incoming data stream $X^*$, called $Y^*$, where each element $Y_i \in Y^*$ contains the range predictions of model $\mathcal{M}_i$, as shown in Equation \ref{model_prediction_intial},

\begin{equation}
    Y^* = \{Y^*_1, Y^*_2, \cdots, Y^*_l\}.
    \label{model_prediction_intial}
\end{equation}

Once these predictions are generated by each model, a likelihood function shown in Equation \ref{raw_likelihood_eq} is used to determine the closest match between the observed phase difference and a model $\mathcal{M}_i$ in the dictionary. This is given by,

\begin{equation}
\begin{split}
    \mathcal{L}_i = log(p(Y_i^*|X^*)) = -\frac{n}{2}\log_{10} 2\pi - \frac{1}{2} det(K_i(X^*,X^*)) \\ - \frac{1}{2}(Y_i - m_i(X^*))^T K_i(X^*,X^*) (Y_i - m_i(X^*))
    \label{raw_likelihood_eq}
\end{split}
\end{equation}
where, $\mathcal{L}_i$ is the computed log likelihood, $Y_i^*$ is the range predictions from the model $\mathcal{M}_i$ using $X^{*}$ and $n$ is the length of the input vector. $Y_i^*$ are generated by a GP Model, thus skewing the results when Equation \ref{raw_likelihood_eq} is applied to the $\mathcal{X}$ and $\mathcal{Y}$ pairs. Therefore, the incoming signal is segmented into three regions (Equation \ref{sorted_data_vec}), where $v
-i$ corresponds to the fraction of the data within the linear region ($v_1$), and outside, $v_2$ and $v_3$. This vector is given by:

\begin{equation}
    \mathbf{v} = [v_1,v_2,v_3]^T
    \label{sorted_data_vec}
\end{equation}
where $v_1$, $v_2$ and $v_3$ are determined using thresholds that are established relatively easily. For this, a linear regression model consisting of range and phase difference pairs, i.e., set of $X$ and $Y$ ordered pairs based on known physical relationships and system properties, is used such as one shown in Equation \ref{linear_predict}, 
  \begin{equation}
    X_{sys} = f(Y) + \sigma_s^2 + x_{sys}
    \label{linear_predict}
\end{equation}
where, $x_{sys}$ are reader offsets that work with the system's standard deviation ($\sigma_s$) to generate $X_{sys}$ based on known $Y$ values. The thresholds are given by Equation \ref{sorting_data_definition},

\begin{equation}
\begin{aligned}
v_1 &=\frac{|\{x[i] \in X^* : \min X_{sys} \leq x[i] \leq \max X_{sys}\}|}{|X^*|} \times 100\% \\
v_2 &= \frac{|\{x[i] \in X^* : x[i] \leq \min X_{sys}\}|}{|X^*|} \times 100\% \\
v_3 &= \frac{|\{x[i] \in X^* : \max X_{sys} \leq x[i]\}|}{|X^*|} \times 100\%. \\
\end{aligned}
\label{sorting_data_definition}
\end{equation}

This process of data segmentation is utilized for each of the training sets to create a set of numerical vectors $v_l$ that describe the distribution of the data associated with each model $M_l$, and with the input data $X^*$ to create a vector $v^*$ that containing its data distribution. Once all of the data sets is sorted, the new data distribution vector $v^*$ is compared with the data distribution of the already studied environments $v_l$ through an Euclidean norm to calculate weights that represent the distance between the data distributions between any two environments, where the weight ($w_l$) is smaller as $v^*$ is closer to $v_i$. The weighted likelihood equation is as shown in Equation \ref{env_likelihood}:

\begin{equation}
\begin{aligned}
 w_l &= \| v_l - v^*\|,\\
Model &= \arg\max_l [w_1 \mathcal{L}_1, w_2 \mathcal{L}_2 \ldots w_l \mathcal{L}_l].
\end{aligned}
\label{env_likelihood}
\end{equation}

 \section{Experiments}\label{implementation}
 This section describes the experiments, details of the RFID system, and the data collection process.

\subsection{Environments and data collection} \label{env_attribute}

\begin{figure*}[htbp]
    \centering
    \begin{subfigure}[t]{0.3\textwidth}
        \centering
        \includegraphics[width=\textwidth]{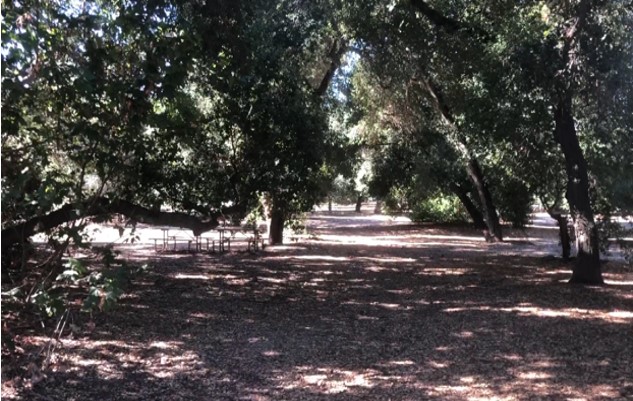}
        \caption{Environment 1}
        \label{fig:env_1}
    \end{subfigure}
    \begin{subfigure}[t]{0.3\textwidth}
        \centering
        \includegraphics[width=\textwidth]{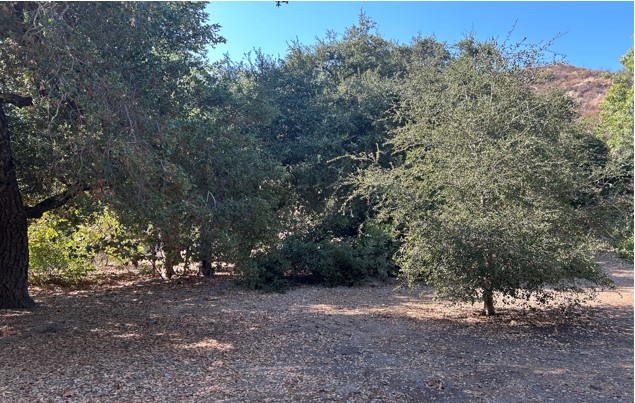}
        \caption{Environment 2}
        \label{fig:env_2}
    \end{subfigure}
    \begin{subfigure}[t]{0.3\textwidth}
        \centering
        \includegraphics[width=\textwidth]{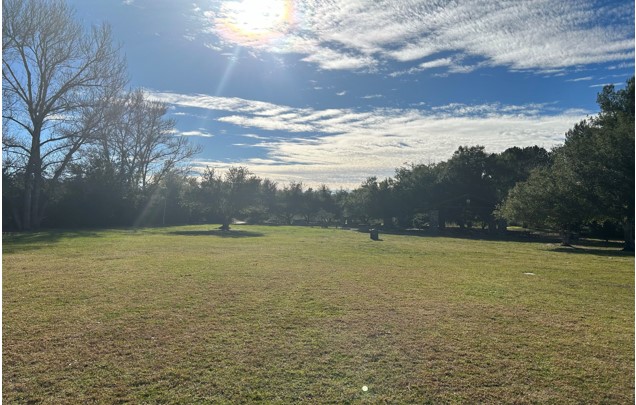}
        \caption{Environment 3}
        \label{fig:env_3}
    \end{subfigure}
    \begin{subfigure}[t]{0.3\textwidth}
        \centering
        \includegraphics[width=\textwidth]{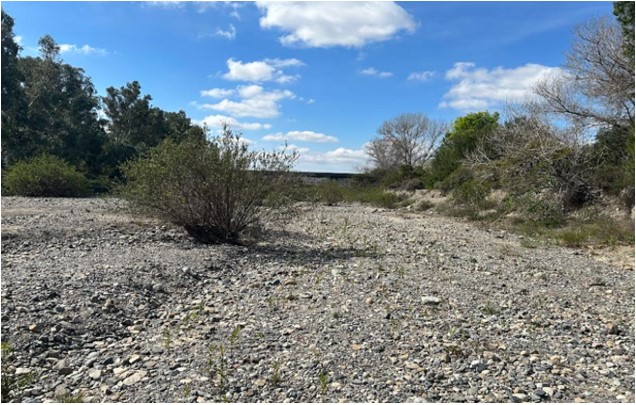}
        \caption{Environment 4}
        \label{fig:env_4}
    \end{subfigure}
    \begin{subfigure}[t]{0.3\textwidth}
        \centering
        \includegraphics[width=\textwidth]{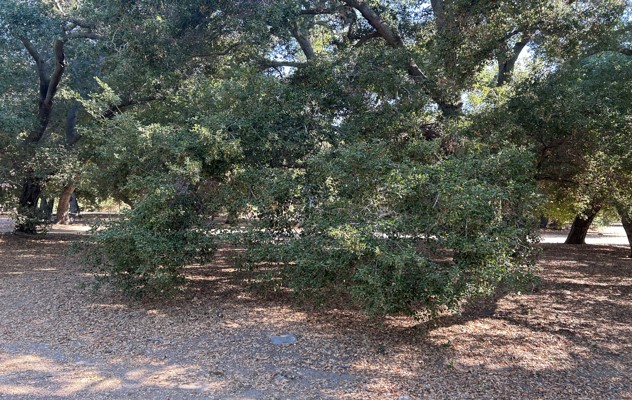}
        \caption{Environment 5}
        \label{fig:env_4}
    \end{subfigure}
    \begin{subfigure}[t]{0.296\textwidth}
        \centering
        \includegraphics[width=\textwidth]{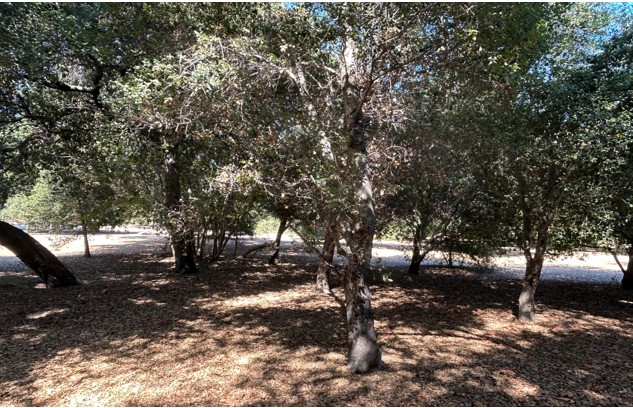}
        \caption{Environment 6}
        \label{fig:env_4}
    \end{subfigure}
    \caption{The six environments used for data collection: Environments 1, 2, and 5 are forested environments that were 100m - 200m away from each other, with Environment 6 nearly half a kilometer away from the other environments. Environments 3 and 4 have minimal obstructions and are used as baselines for grassy and rocky terrain.}
    \label{fig:environments}
    \vspace{-11pt}
\end{figure*}

 For this paper, the dictionary includes six types of park environments: four forest areas, one grassland, and a semi-arid region characterized by bushes and rocky materials. These environments represent the typical areas where fires occur and spread, based on historical data. Each environment will be labeled as Environments 1 through 6, and sample images from these environments are presented in Figure \ref{fig:environments}. To demonstrate the adaptability of the generated GP Model, each environment is used as a calibration baseline for the other environments, to demonstrate the accuracy of the system the correct model is used.

\subsubsection{Environmental Mapping with Single Tag}

\begin{figure}[htbp]
    \centering
    \begin{subfigure}[b]{0.23\textwidth}
        \centering
        \includegraphics[width=\textwidth]{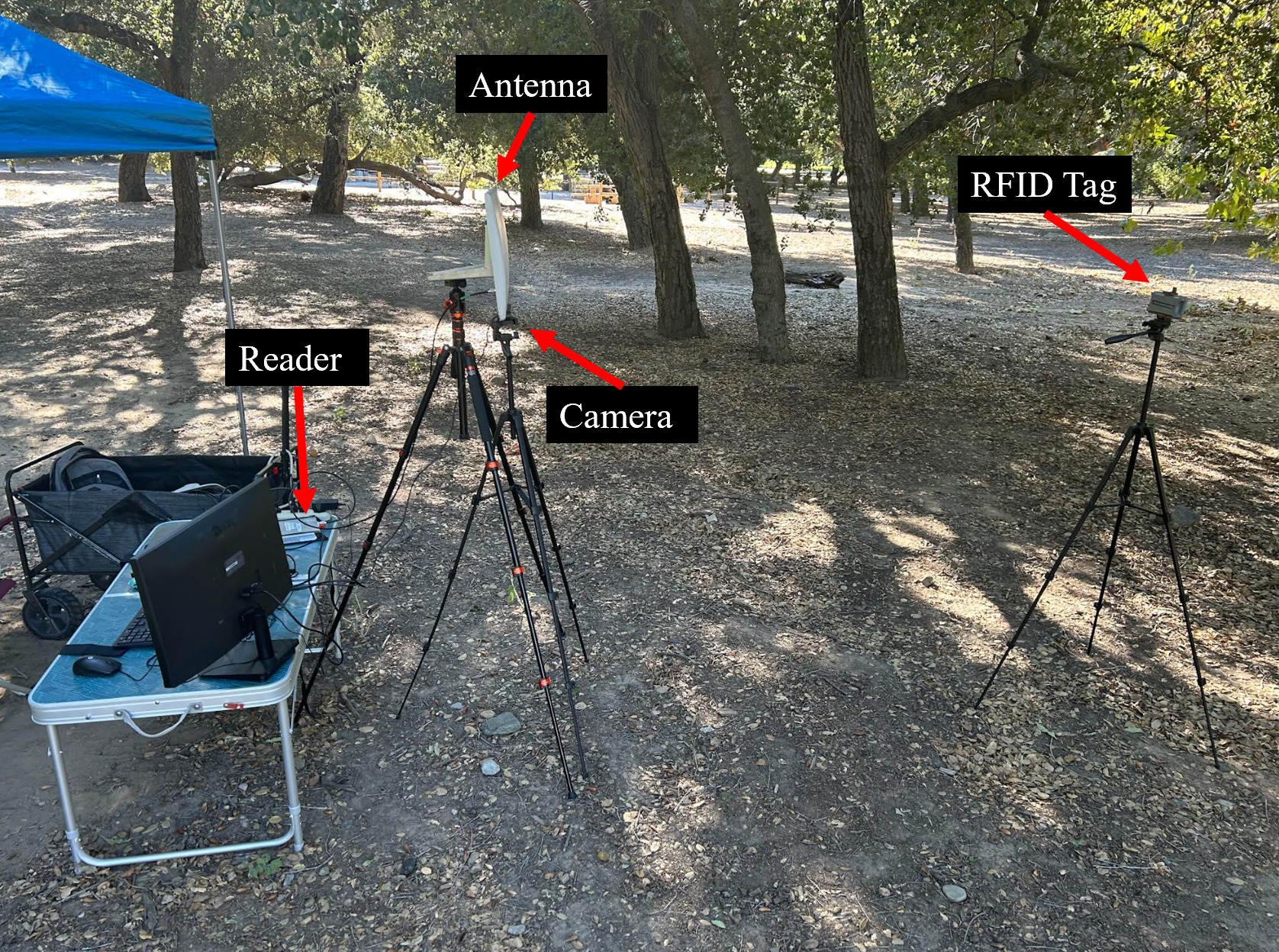}  
        \caption{}
        \label{tag_setup}
    \end{subfigure}
    \begin{subfigure}[b]{0.23\textwidth}
        \centering
        \includegraphics[width=\textwidth]{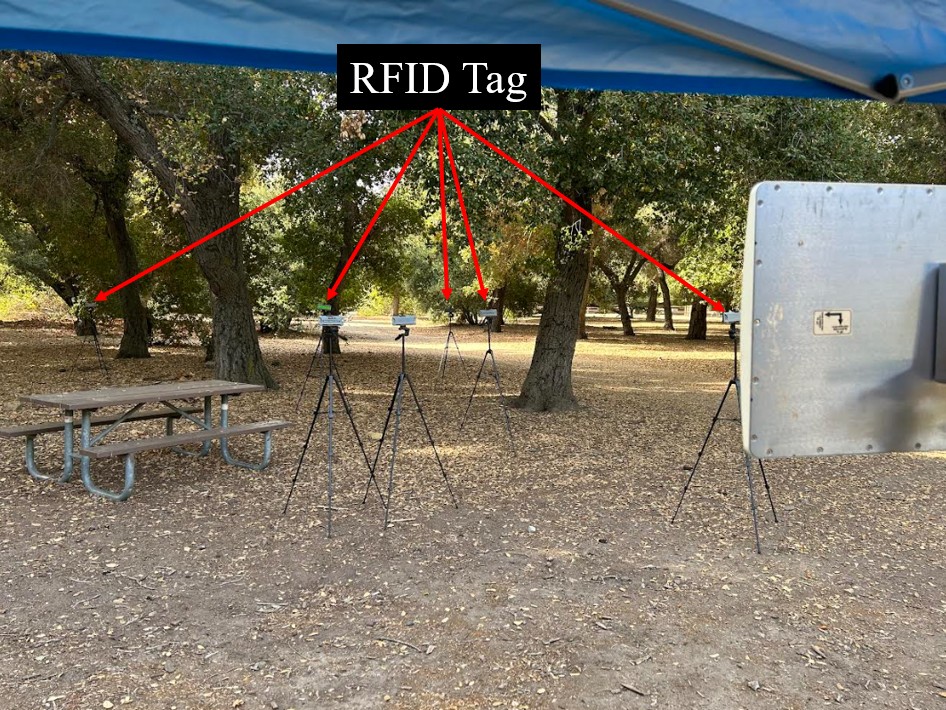}  
        \caption{}
        \label{multi_tag_setup}
    \end{subfigure}
    \caption{Data was collected by placing the tag at varying distances and angles from the antenna centerline, using a camera and measuring tape to record positions. Over 15 days, approximately 27,000 data points were gathered per environment, totaling 162,000 across six environments. While mapping signal attributes for a single tag, multiple tags were also placed to assess system performance under multi-tag conditions. Although RFID systems activate one tag at a time due to anti-collision protocols, the physical presence of other tags can still interfere with the signal.}
    \vspace{-18pt}
    \label{data_collection}
\end{figure}

Data was collected manually by placing an RFID tag 2–20 meters from the reader in 0.5-meter increments, and across angles from $-30^\circ$ to $30^\circ$ in $15^\circ$ steps. At each position, data was sampled at 50 Hz for 3–5 minutes, resulting in about 200 samples per position and 135 positions per environment. RSSI was measured at 902.75 MHz, and phase change was recorded between 902.75 and 903.75 MHz—selected for optimal detection after testing various frequencies. Data was split 80/20 for training and evaluation. Tag distances were measured using a tape measure and a camera tracking a team member next to the tag, with the camera only providing data up to 10 meters, its effective range. The setup is shown in Figure \ref{tag_setup}.

\subsubsection{Multiple Tag Experiments}

To evaluate multi-tag performance, 8 tags were deployed in Environment 1 and repositioned across 15 trials to simulate mobility. Signal attributes were recorded for 5 minutes per trial, with the reader individually querying each tag via standard anti-collision protocols \cite{rfid_protocol_3}. Over 12 additional trials, tags were moved to new positions and orientations—including close proximity and occlusion—to assess signal interference. Although limited to 8 tags due to setup constraints, the system can scale based on processing speed, environmental complexity, reader capabilities, and update frequency. The Impinj R220 Reader used supports up to 200 tag reads per second.
 
 \subsection{System Hardware}
 
The RFID system used EXO 3000 passive tags, an MTI MT-263020 (RHCP) outdoor antenna, and an Impinj R220 reader, with no hardware or firmware modifications. Operating between 902.75–927.25 MHz (FCC-compliant and outside the 140--155 MHz firefighter radio band), the system functions at temperatures up to 85--100 $^\circ$C. Ground truth distances were measured using a tape measure and a Zed 2 camera. RFID data was stored on a Dell Inspiron 14 laptop, and camera data on a dedicated Zed Box. Data was merged in Excel and processed using C$\#$, Python, and MATLAB. While data was collected using the Zed Box and laptop, MATLAB code ran on a desktop with a 3.00GHz Intel Core i7-9700F and 16GB RAM. The reader connected to the laptop via Ethernet; the camera connected to the Zed Box via USB.

 \section{Results and Discussion}\label{results}


This section is organized as follows: first, we discuss how the environment affects key RFID signal features. Next, we analyze how kernel and model choices influence range prediction accuracy. Then, we compare our results with the K-Nearest Neighbor algorithm \cite{KNN_1,KNN_2} and a double-layer Deep GP model \cite{GP_flux}, highlighting the ability of GP models to predict sensor data using both correct and incorrect training sets. Finally, we demonstrate the effectiveness of likelihood estimation and the importance of weighted likelihood for selecting the correct model.

 \begin{figure}[htbp]
    \centering
    \begin{subfigure}[b]{0.47\textwidth}
        \centering
        \includegraphics[width=\textwidth]{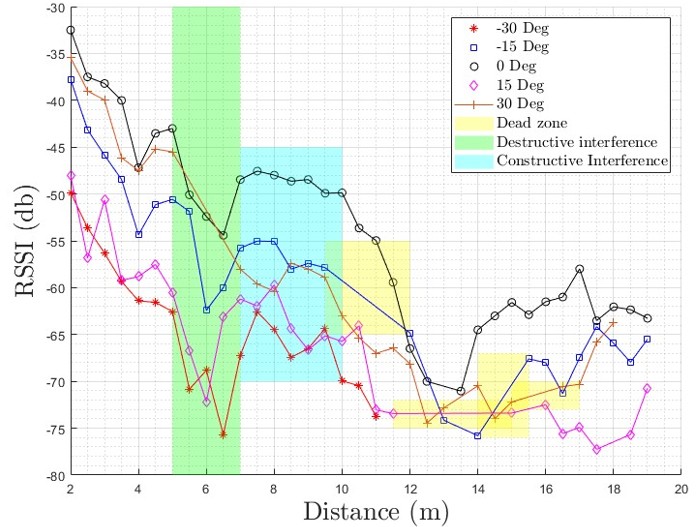}  
        \caption{Env 2 RSSI across all angles}
        \label{env2_rssi}
    \end{subfigure}
    \hfill
    \begin{subfigure}[b]{0.47\textwidth}
        \centering
        \includegraphics[width=\textwidth]{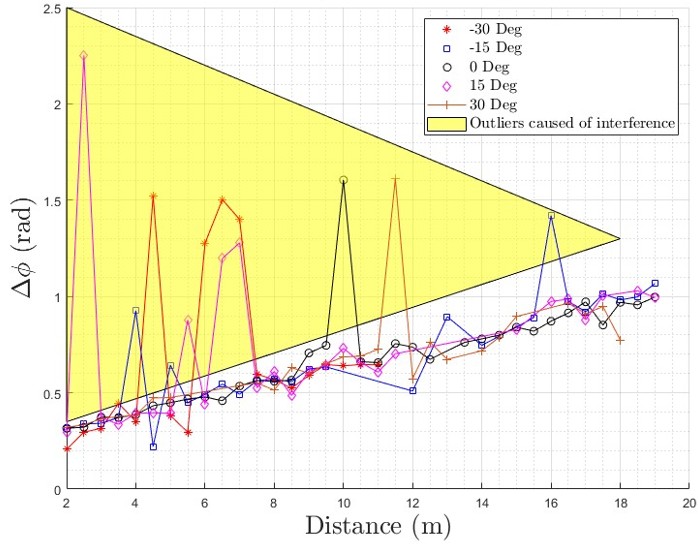}  
        \caption{Env 2 $\Delta\phi$ across all angles}
        \label{env2_phase}
    \end{subfigure}
    \caption{This figure shows the signal attributes observed in Environment 2. Although each environment had its own unique signal response, they all shared a general trend in signal behavior. Each environment exhibited distinct construction and destructive interference regions, including dead zones, which manifested as outliers in the $\Delta\phi$ data.}
    \vspace{-11pt}
    \label{env2_signal_bevahior}
\end{figure}

\subsection{Environmental Influence on RFID Signal} \label{ev_influence}
Localizing sensors in outdoor forest environments is challenging due to environmental variability and limited control, even within small areas. Although brush density may appear consistent, factors like brush type, moisture, tree structure, and animal presence can significantly impact signal propagation. This was reflected in our data, where tags were sometimes undetectable or experienced severe signal interference, evident as large spikes in phase and RSSI values (see Figure \ref{env2_signal_bevahior}). While Figure \ref{env2_signal_bevahior} shows data from Environment 2, similar behavior occurred across all environments at varying tag positions. These unpredictable RF disturbances introduced two core challenges that had to be addressed for reliable outdoor operation.

The first challenge was the effect of foliage and terrain on signal distribution. In certain locations, destructive interference caused significant drops in RSSI and large fluctuations in phase difference ($\Delta\phi$), as shown in Figure \ref{env2_signal_bevahior}. Peaks in $\Delta\phi$ (highlighted in yellow in Figure \ref{env2_phase}) often aligned with RSSI troughs (green in Figure \ref{env2_rssi}). These disruptions varied unpredictably by position and angle, complicating signal mapping and localization.

The second challenge involved “dead zones”—areas where tags were undetectable. Figure \ref{env2_rssi} shows that some tag orientations failed to register, while mirrored positions at the same distance were successful. A slight position change often restored detection, indicating these dead zones result from localized destructive interference shaped by terrain and foliage.

Although signal drops hinder detection, they often follow predictable patterns. As tags near dead zones, RSSI declines rapidly before contact is lost. To handle this, a detection algorithm—such as those in \cite{tag_loss_1}—can monitor RSSI and read rate to flag potential dead zones and apply safeguards, like constraints in Equation \ref{tag optimization eq}, to preserve localization accuracy.

However, these safeguards may fail when tags move too far from the antenna centerline. At angles like $30^\circ$ and $-30^\circ$ (Figure \ref{env2_rssi}), the signal can abruptly vanish without warning. This highlights the need for more adaptive precautions, especially near the edge of the antenna’s detection cone, where generic safeguards may be insufficient.

\begin{figure*}[htbp]
    \centering
    \begin{subfigure}[b]{0.3\textwidth}
        \centering
        \includegraphics[width=\textwidth]{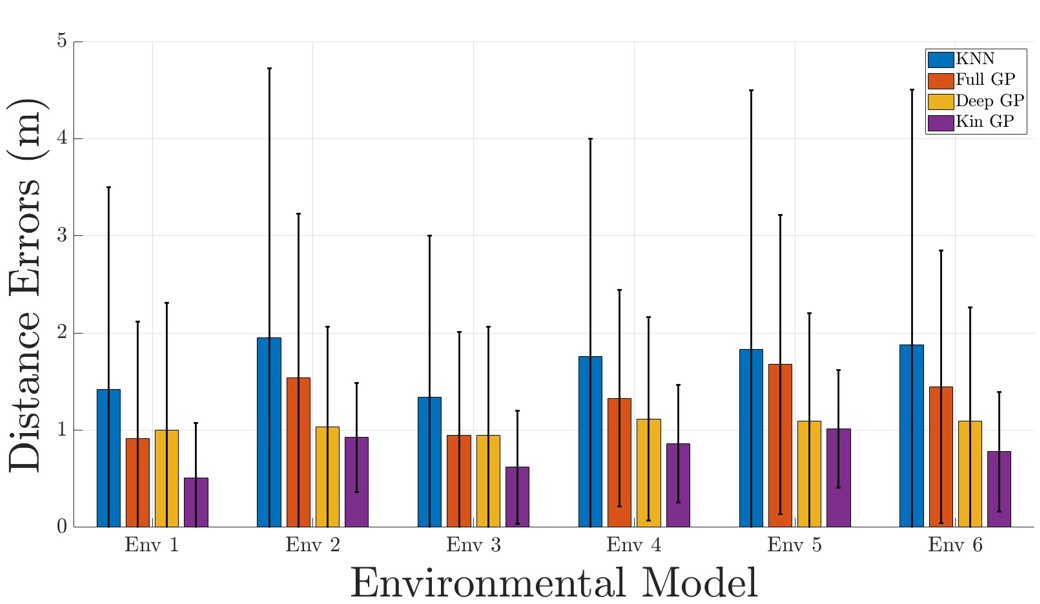}
        \caption{Env 1 Mean Distance Errors}
        \label{m1_e1}
    \end{subfigure}
    \hfill
    \begin{subfigure}[b]{0.3\textwidth}
        \centering
        \includegraphics[width=\textwidth]{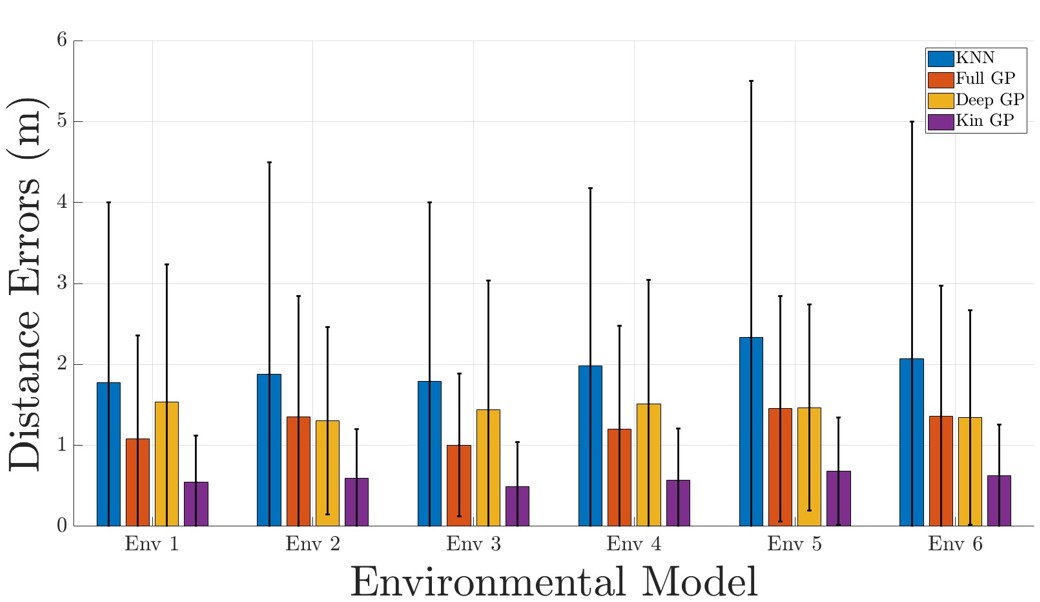}
        \caption{Env 2 Mean Distance Errors}
        \label{m1_e2}
    \end{subfigure}
    \hfill
    \begin{subfigure}[b]{0.3\textwidth}
        \centering
        \includegraphics[width=\textwidth]{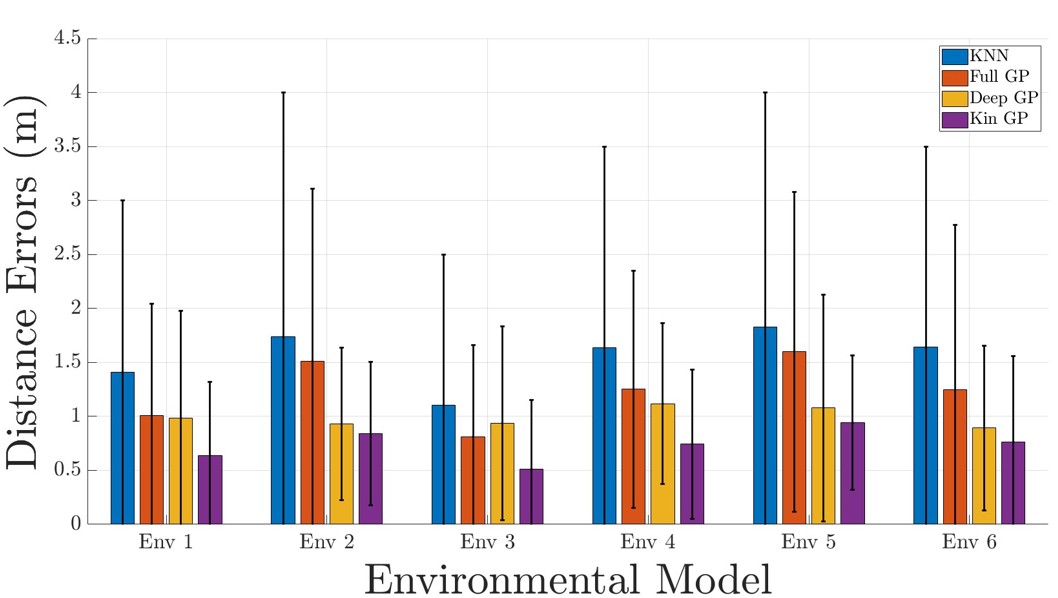}
        \caption{Env 3 Mean Distance Errors}
        \label{m1_e3}
    \end{subfigure}
    
    \vspace{0.5em}  

    \begin{subfigure}[b]{0.3\textwidth}
        \centering
        \includegraphics[width=\textwidth]{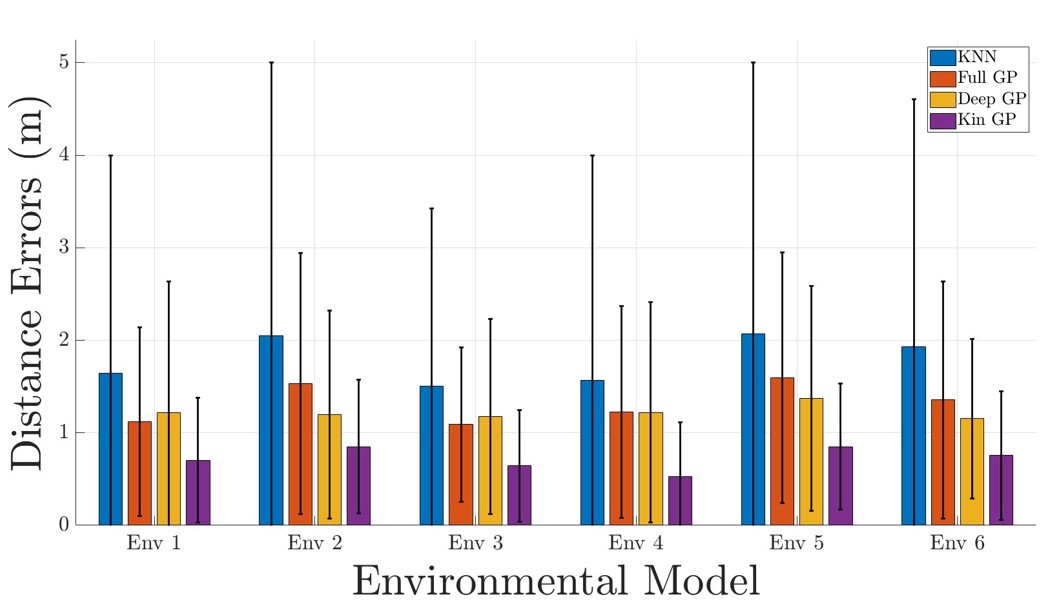}
        \caption{Env 4 Mean Distance Errors}
        \label{m1_e4}
    \end{subfigure}
    \hfill
    \begin{subfigure}[b]{0.3\textwidth}
        \centering
        \includegraphics[width=\textwidth]{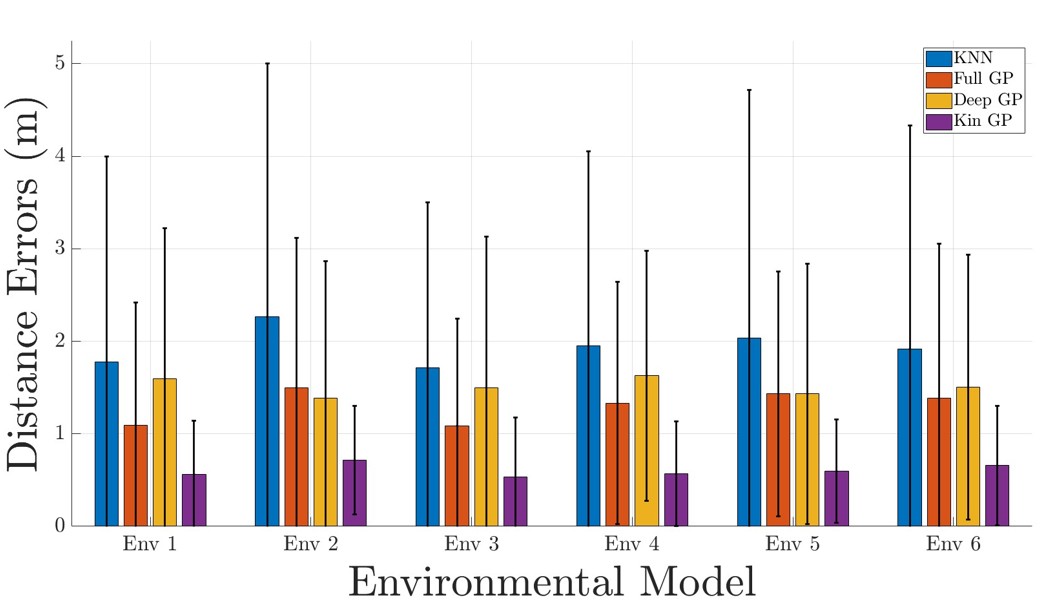}
        \caption{Env 5 Mean Distance Errors}
        \label{m1_e5}
    \end{subfigure}
    \hfill
    \begin{subfigure}[b]{0.3\textwidth}
        \centering
        \includegraphics[width=\textwidth]{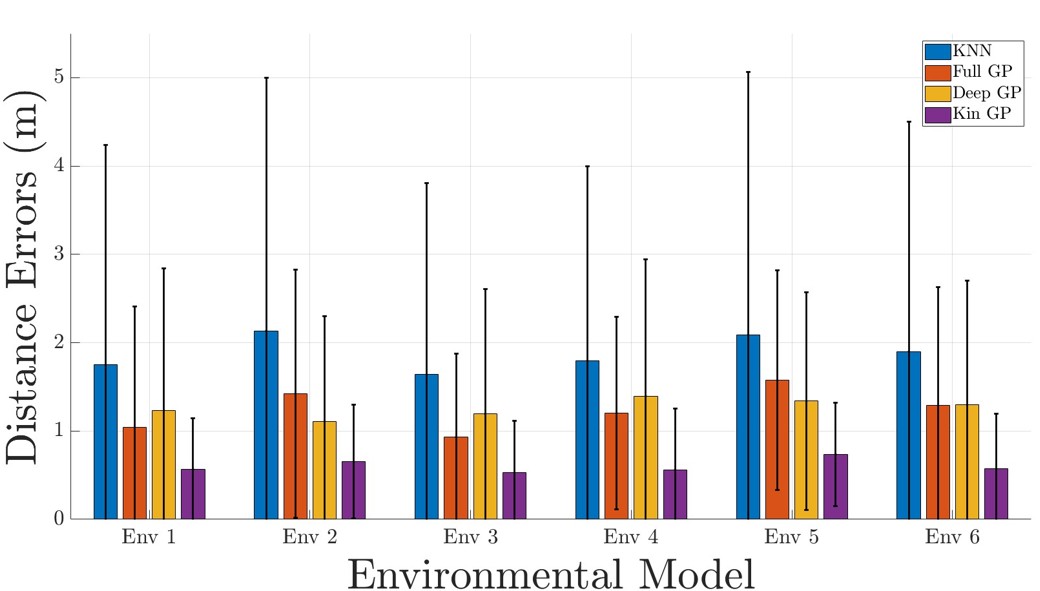}
        \caption{Env 6 Mean Distance Errors}
        \label{m1_e6}
    \end{subfigure}
    \caption{This figure illustrates the mean distance errors for each environment and the corresponding 90th percentile error ranges generated by the KNN and both GP methods. The GP outperforms KNN, and by incorporating information about the tag's trajectory, the accuracy can be further improved, achieving a range error of less than 2.5 m in most cases, a full meter below the errors obtained through the application of KNN.}
    \vspace{-11pt}
    \label{error_bars}
\end{figure*}

\subsection{Application of GP Model}\label{GP app}

In this section, we first present the results and discuss the impact of the model by comparing the resulting distance mean error values and their associated 90th percentile rages. Next, we analyze the effect of kernel choice on these errors by comparing the Radial Basis Function used in this study with the Lagrange Kernel, Matern Kernel, and Quadratic Kernel, all of which utilize the proximity between points to generate covariance matrices. The length hyperparameter for the Gaussian kernel was set to 0.0075, a value determined through a standard tuning process and maintained as constant throughout the study.

\subsubsection{Effect of Model Choice}\label{model_choice}

Figure \ref{error_bars} illustrates the calculated distance uncertainties when test data for each environment was evaluated using the environmental models generated through GPs, a double layer Deep GP model and the KNN algorithm . All model types delivered range estimates comparable to those produced by commercially available GPS devices. However, the GP models demonstrated superior accuracy, achieving errors that were 25\% to 50\% lower than those produced by the KNN algorithm. Additionally, the mean ranging errors when using KNN fluctuated between 1.2 to 2.4 meters with the maximum ranging errors fluctuating between 5 to 6 meters, depending on the selection of the model. In contrast, the GP models provided more consistent results, with mean errors ranging from 0.8 to 2.25 meters and maximum errors less than 3.5 meters depending on which GP model was applied to the incoming data stream. When comparing a base GP model and a deep GP model, it is evident that a deep GP Model is able to provide on average lower localization error than base GP Models. However, the order of magnitude of the errors can be measured in the 10s of centimeter range, which is negligible in comparison to the 10s of meter range in which the system operates. Furthermore, by applying a kinematic constraint of $\pm~2$ meters—based on the maximum movement speed of fire fighters in the field and potential equipment errors—the GP models regularly achieved sub-meter level ranging errors, regardless of the model used. These results highlight that while the KNN algorithm can compete with GPS regarding positioning errors, the GP models consistently perform better even if an inappropriate model is utilized for the environment.

 \subsubsection{Effect of Kernel Choice on System Accuracy}
 
One important factor to consider when applying a GP model is how different kernels affect the system, as each kernel represents a unique relationship metric between data points. To illustrate the impact of various kernels, we compared the root mean square error (RMSE) values when using the Lagrangian Kernel, Matern Kernel, Quadratic Kernel, and Radial Basis Function (RBF) Kernel. This comparison was conducted on Environment 4, utilizing itself and Environment 6 as training data to demonstrate both a correct and incorrect environmental match. The results of this analysis are shown in Table \ref{different_kernel}, which indicates that the Radial Basis Function provided the best performance among the tested kernels. This reinforces the well-known issue in GP models regarding the significant influence of kernel choice on the outcomes and the need for an appropriate choice.
\begin{table}[htbp]
    \centering
    \begin{tabularx}{\linewidth}{|X|c|c|}
        \hline
        \textbf{Kernel Type} & \textbf{Correct Env RMSE (m)} & \textbf{Incorrect Env RMSE (m)} \\
        \hline
        $K_{Matern}$ & 3.33 & 3.53 \\
        \hline
        $K_{Lagrange}$ & 3.43 & 3.43 \\
        \hline
        $K_{Quadratic}$ & 5.70 & 3.53 \\
        \hline        
        $K_{RBF}$ & 1.06 & 1.57 \\
        \hline
    \end{tabularx}        
    \caption{Despite being tuned to provide the best estimates, the other kernels performed poorly compared to the Radial Basis Function. The ranging errors are approximately 3 times larger than those produced using the Radial Basis Function, thus underscoring the well-known issue of GP, namely the need for proper kernel choice.}
    \vspace{-11pt}
    \label{different_kernel}
\end{table}
\begin{figure}[htbp]
    \centering
    \includegraphics[width=0.47\textwidth]{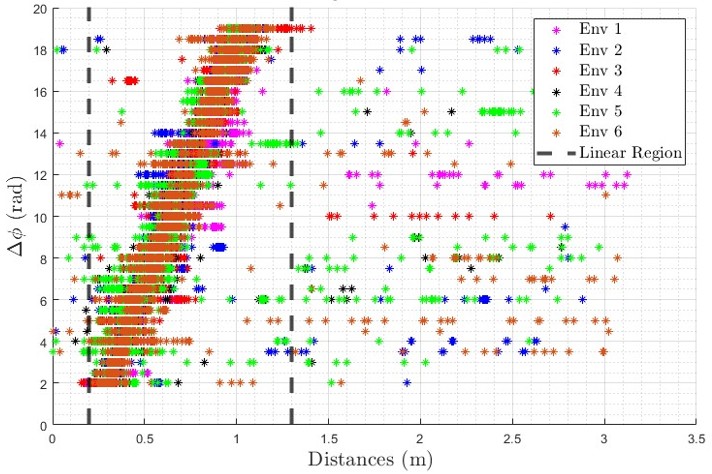}  
    \caption{Without prior range information, thresholds are established using the phase values to delineate three regions, providing a basis for the weighted likelihood function.}
    \vspace{-11pt}
    \label{data_segmentation}
\end{figure}

\subsection{GP Model Selection from a Dictionary}

As demonstrated previously in Section \ref{model_choice}, identifying the correct environment using the RF signal response produces more accurate range measurements, which is further improved through kinematic corrections. To explore model selection through weighted likelihood, 100 trials were performed with each trial featuring different permutations of the data set, treating it as a sparse data set by only using a portion of the available phase data, as shown in Figure \ref{data_segmentation}.

Figure \ref{data_segmentation} shows the phase-range relationship for all the environments. Also shown on this figure are thresholds used to define the linear range of data, with $\Delta\phi_{linear} \in [0.2,1.2]$ radians. Table~\ref{env_weights} shows the distribution of data that falls in each region corresponding to $v_1,~v_2,~v_3$ in Equation \ref{sorting_data_definition}. $v_2$ shows the fraction of the data that falls in the linear region, which typically contains more than 94\% of the data in all cases except Environment 3. Environment 3 was the most dissimilar from a RF response standpoint from the remaining environments and this was supported in the results with this environment correctly classified using the ML approach in nearly all the trials. Environment 1 was also classified correctly in nearly all the trails as its signal response was the most linear among all the environments. The remaining environments were correctly classified using the ML approach with the exception of Environment 2 which was frequently classified as Environment 1. Overall, the weighted ML method performed well, particularly when the RF response of the environments were dissimilar.

\begin{table}[htbp]
    \centering
    \begin{tabular}{|c|c|c|c|c|c|c|}
        \hline
         \textbf{Fraction \%} & \textbf{Env 1} & \textbf{Env 2} & \textbf{Env 3} & \textbf{Env 4}  & \textbf{Env 5}  & \textbf{Env 6}\\ 
         \hline
         $v_1$ & 0.1 & 1.4 & 2.8 & .2 & 0.6 & 0.5\\        
         \hline
         $v_2$ & 98.7 & 94.2 & 82.9 & 96.0 & 93.3 & 95.9 \\
         \hline
         $v_3$  & 1.2 & 4.5 & 14.4 & 3.8 & 6.1 & 3.6\\        
          \hline
    \end{tabular}
    \caption{Table showing the data fractions generated for each of the environments in the three ranges using Equation \ref{sorting_data_definition}.}
    \label{env_weights}
\end{table}

\subsection{Effect of Multiple Tags on Range Accuracy}

Due to built in anti-collision protocols, RFID readers can only communicate with one RFID tag at a time, which prevents the radio signal from being influenced by communication between the reader and  other RFID tags \cite{rfid_protocol_3}. Therefore, any additional tags dispersed in the environment can only act as a physical obstruction to the radio signal itself. In addition, the multi-tag experiments were performed several months after Environment 1 had been initially mapped. During this time, the environment changed with the depletion of vegetation and the addition of objects such as picnic tables. Therefore, when the GP Model is applied to the new tag positions, the average range mean error values between single tag and multiple tag experiments changed, as shown in Table \ref{multi_tag_complete_response}.

\begin{table}[htbp]
    \centering
    \begin{tabular}{|c|c|c|c|c|}
        \hline
         \textbf{Model } & \textbf{Env Used} & \textbf{MT ME (m)} & \textbf{ST ME(m)} & \textbf{MT Z-Score}\\ 
         \hline
         Full Data & Env 1 & 1.83 & 0.91 & 0.92\\        
         \hline
         Kin Cst & Env 1 & 0.60 & 0.51 & 0.22 \\
         \hline
         Full Data  & Env 2 & 2.17 & 1.54 & 0.31\\        
          \hline
         Kin Cst & Env 2 & 0.55 & 0.92 & -0.26\\
         \hline
    \end{tabular}
    \caption{The mean error values that were obtained by applying the GP Models to each of the deployed RFID tags compared to when only one tag was deployed.}
    \label{multi_tag_complete_response}
\end{table}

Despite such physical changes in the environment resulting from the period elapsed between the two experiments, results in Table \ref{multi_tag_complete_response} show that when the entire dataset is used, the mean error values when multiple tags were deployed is within one standard deviation to the mean values calculated when there is only one tag present in the environment. This same behavior was seen when the kinematic constrains were applied with the system routinely achieving submeter level accuracy. Therefore it is safe to conclude that despite changes that occurred to the environment in a multiple-tag setting, the system provides GPS level localization accuracy, further supporting its use in wildfire applications. 

\begin{figure*}[htbp]
    \centering
    \begin{subfigure}[t]{0.30\textwidth}
        \centering
        \includegraphics[width=\textwidth]{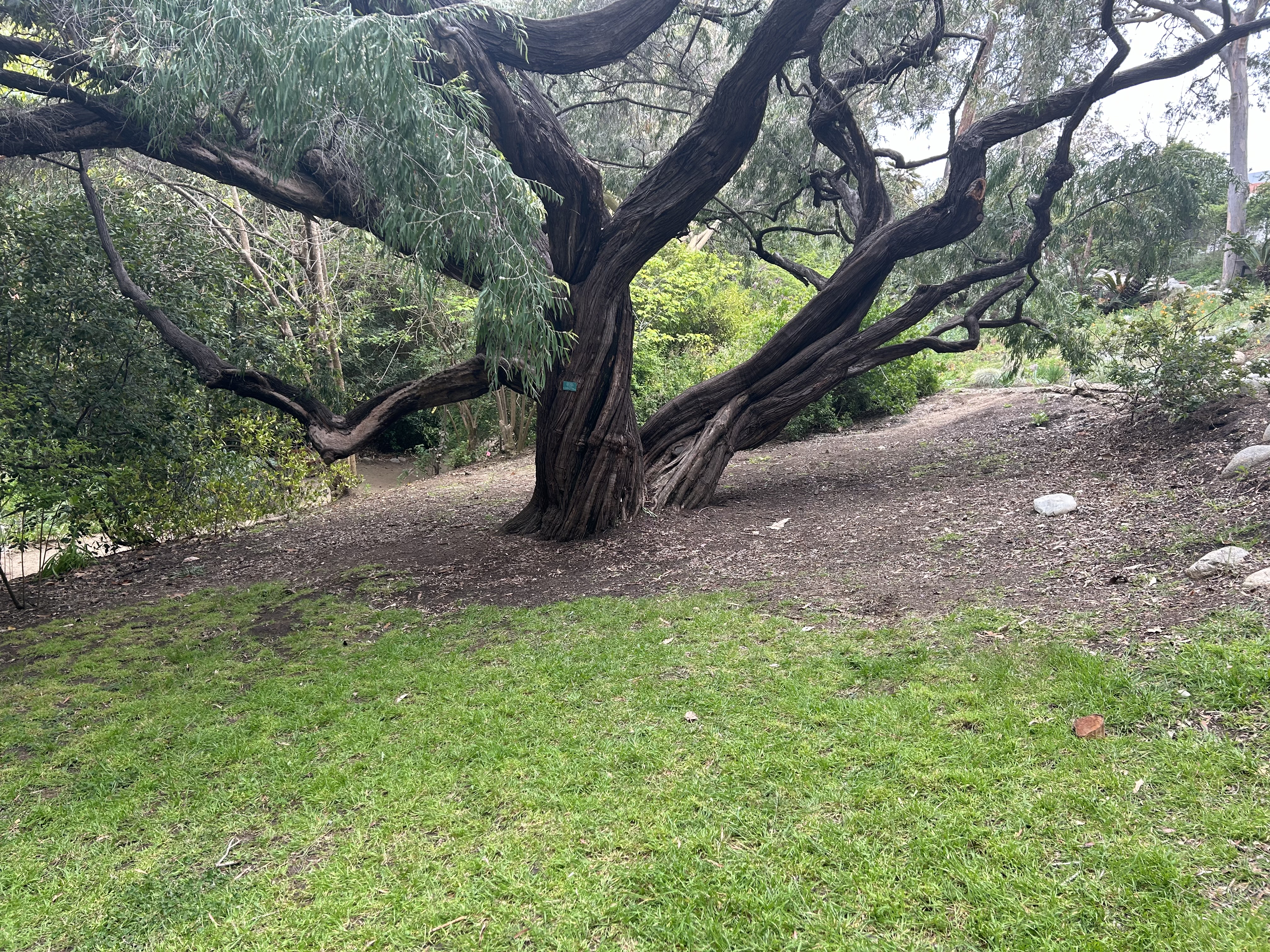}
        \caption{Env A}
        \label{fig:env_1}
    \end{subfigure}
    \begin{subfigure}[t]{0.30\textwidth}
        \centering
        \includegraphics[width=\textwidth]{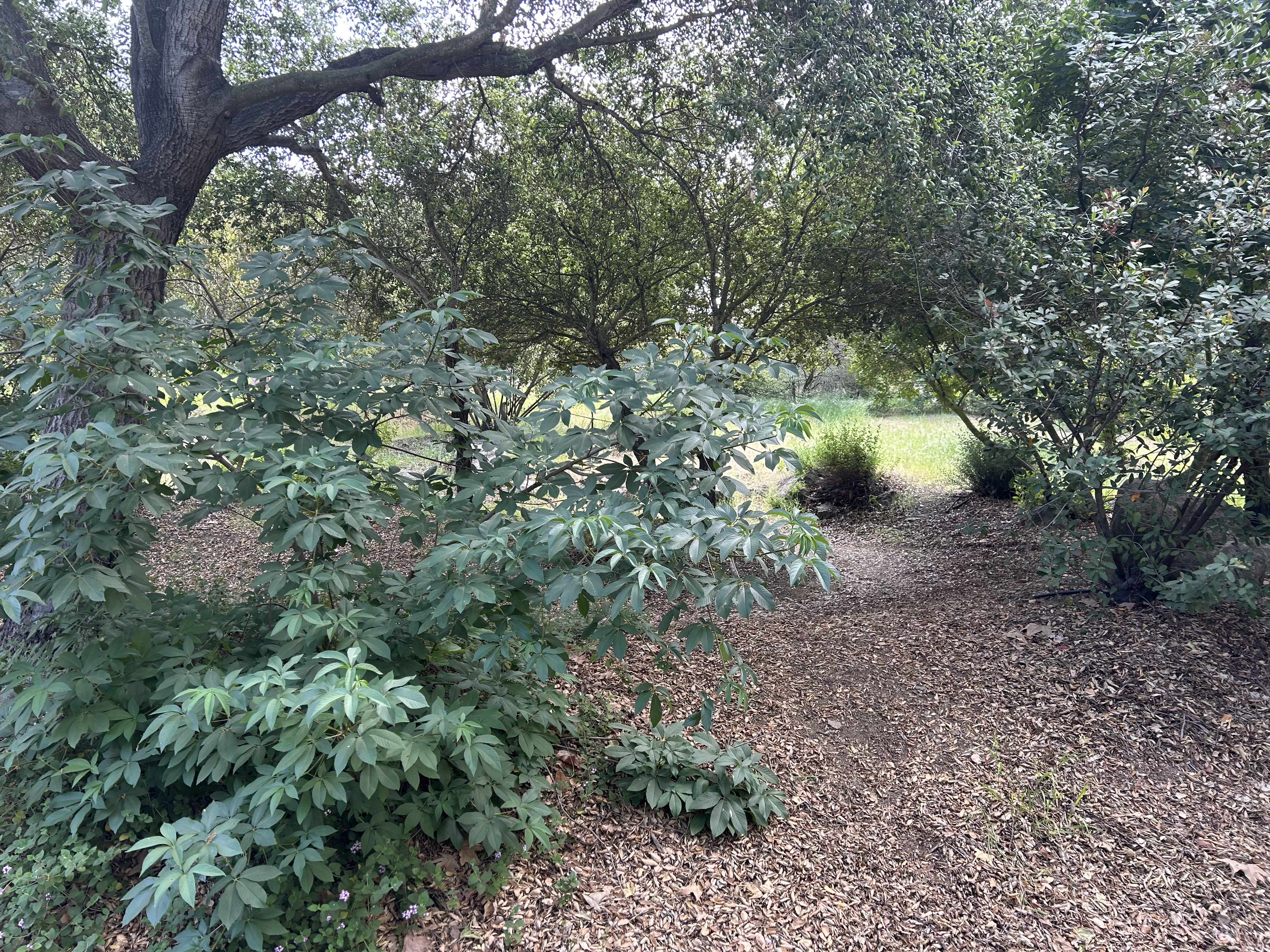}
        \caption{Env B}
        \label{fig:env_2}
    \end{subfigure}
    \begin{subfigure}[t]{0.30\textwidth}
        \centering
        \includegraphics[width=\textwidth]{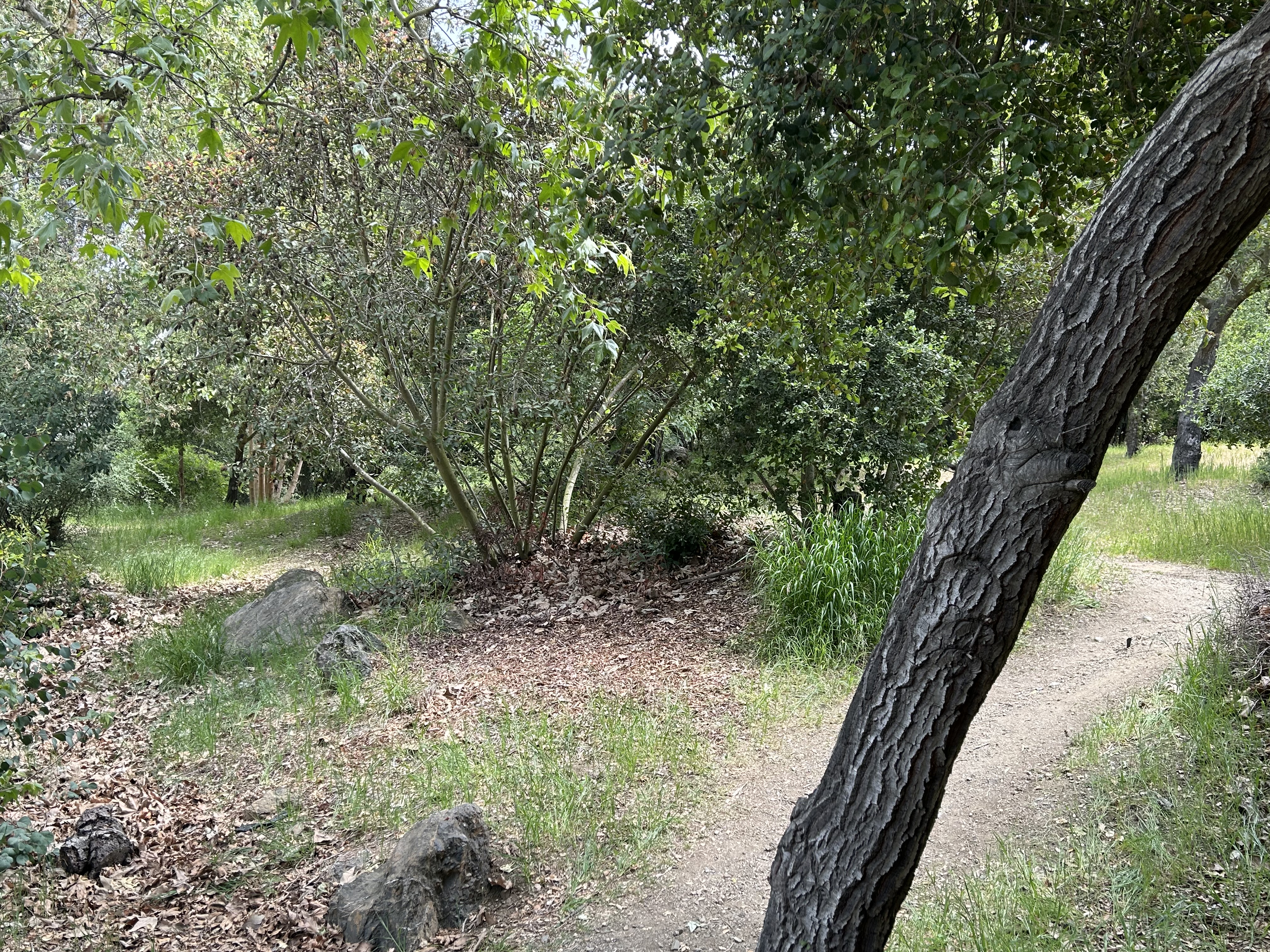}
        \caption{Env C}
        \label{fig:env_3}
    \end{subfigure}    
    \caption{The approach was tested in three previously unmeasured environments to determine how the system would respond. These three new environments have sloping terrain, large boulders, and wet foliage that were all absent from previously modeled environments.}
    \label{new_env}
    \vspace{-11pt}
\end{figure*}

\subsection{Results from Unmeasured Environments}\label{new_multi_tag_env}

From a practical standpoint, it is not possible to include all environment types in the environmental dictionary. A key advantage of this new approach is its potential to improve ranging accuracy even while operating in a previously unmeasured environment. To demonstrate this, multiple tags were dispersed in three new environments --A, B, C -- located over 100 miles away from Environments 1 and 2, and contain sloped terrain, taller trees, denser foliage, large boulders and wet terrain due to recent rains that were not present in the original environmental dictionary, and are shown in Figure \ref{new_env}. For these experiments, Environments 1 and 2 were used as the training set since they represented forested environments with low and high noise levels, respectively. The average mean errors for the tags in the environments are displayed in Table \ref{multi_env_table}. Results from these experiments show that the method yields range estimates comparable to GPS performance in these terrain conditions, even under these significant differences in the topography between environments.

\begin{table}[htbp]
    \centering
    \begin{tabular}{|c|c|c|c|c|}
        \hline
         \textbf{Env} & \textbf{Env 1 Full} & \textbf{Env 1 Kin} & \textbf{Env 2 Full} & \textbf{Env 2 Kin} \\ 
         \hline
         Env A ME  & 3.36 m & 0.90 m & 2.52 m & 0.57 m \\ 
         \hline
         Env B ME & 3.50 m & 0.78 m & 2.74 m & 0.69 m\\ 
         \hline
         Env C ME & 3.20 m & 0.84 m & 2.85 m & 0.75 m\\
         \hline    
    \end{tabular}
    \caption{To compare the GP Model response in new environments, two measured forest models were used as calibration baselines for the newly measured environments. Despite environmental differences, such as sloped terrain, large boulders and denser foliage, the GP Model was still able to produce range estimates comparable to commercially available GPS devices.}
    \vspace{-11pt}
    \label{multi_env_table}
\end{table}

\section{Conclusion}\label{conc}

To improve asset tracking in active wildfires, this paper presents a novel framework for localizing low-cost passive RFID tags in outdoor environments using a Gaussian Process (GP) model and weighted log-likelihood—a localization scheme that, to the best of the author's knowledge, has not been previously explored. This method achieves distance errors between 0.5 and 3 meters, comparable to GPS accuracy. Coupled with the reader’s capacity to scan up to 200 tags per second and the model’s ability to localize dozens of tags per minute on a consumer-grade laptop, the system shows strong potential for real-world scalability, while still being affordable to low funded fire departments.

However, the method has limitations. The current environmental dictionary includes only six environments within a limited geographic area. Although Section \ref{new_multi_tag_env} demonstrates the model’s adaptability to unseen terrain, broader coverage is needed to improve accuracy and usability. The system also depends on deploying mobile platforms (e.g., robots or personnel) with localization tools like GPS, as passive RFID's range is limited to tens of meters. Scaling to large fires will require multiple mobile readers. Future work will address tracking assets that move out of range and designing search algorithms for lost tags.

On scalability, while performance will vary by application, the system is designed to support clusters of several dozen tags assigned to teams or individuals. Readers—mounted on personnel or vehicles—can operate passively within 20–30m, or up to 100 m with a small battery. Localization inference supports about 30 tags per minute on a consumer PC, which is sufficient for asset tracking at walking speeds (1.2–1.4m/s). Challenges remain, particularly in ensuring continuous coverage and tracking beyond the reader’s range—areas that warrant further research.

\bibliographystyle{IEEEtran}
\bibliography{refrences}

\end{document}